\documentclass[10pt,twocolumn,letterpaper]{article}
\usepackage[pagenumbers]{cvpr}
\definecolor{cvprblue}{rgb}{0.21,0.49,0.74}
\usepackage[pagebackref,breaklinks,colorlinks,allcolors=cvprblue]{hyperref}
\hypersetup{
    pdftitle={Loop4D: Embodied World Modeling with 4D VAEs and Looped Diffusion Transformers},
    pdfauthor={Gen Li, Fan Luo, Mengjun Lin, Shu Han, Bohan Li, Hao Zhao}
}

\title{Loop4D: Embodied World Modeling\\
with 4D VAEs and Looped Diffusion Transformers}

\author{Gen Li\textsuperscript{1,*}\quad Fan Luo\textsuperscript{2,*}\quad Mengjun Lin\textsuperscript{2}\quad
Shu Han\textsuperscript{3}\quad Bohan Li\textsuperscript{4}\quad Hao Zhao\textsuperscript{2}\\[4pt]
{\normalsize \textsuperscript{1}Zhejiang University\qquad \textsuperscript{2}Tsinghua University}\\
{\normalsize \textsuperscript{3}University of Wisconsin - Madison\qquad \textsuperscript{4}Shanghai Jiao Tong University}\\
{\footnotesize \textsuperscript{*}Gen Li and Fan Luo contributed equally to this work.}}

\begin{document}
\makeatletter
\twocolumn[{
    \@maketitle
    \vspace{-14.5pt}
    \begin{minipage}{\textwidth}
\centering
\includegraphics[width=\linewidth]{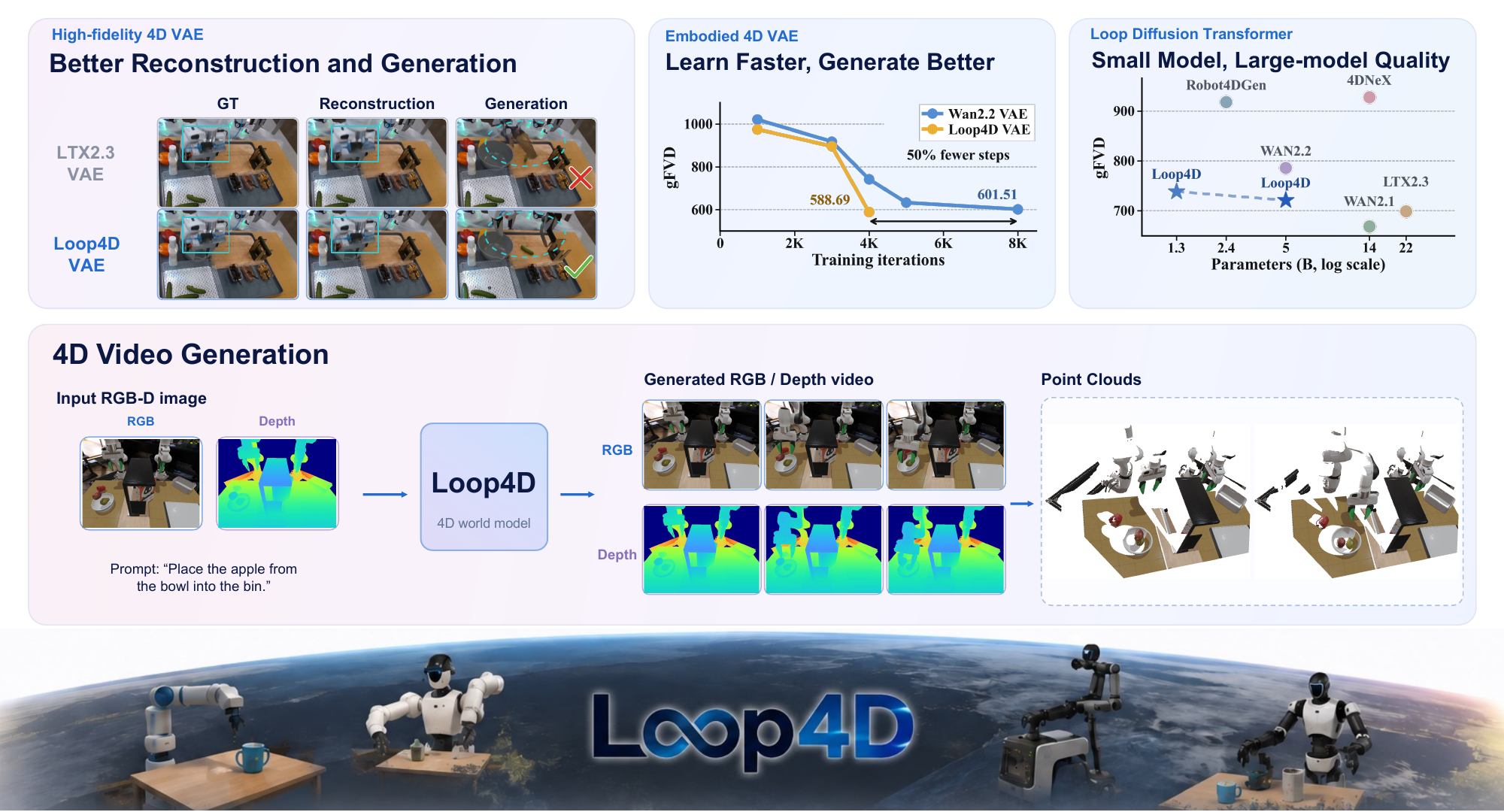}

\captionof{figure}{\textbf{Loop4D: a 4D world model for embodied intelligence.}
(a) Our 4D VAE preserves robot appearance and geometry during motion, enabling high-quality reconstruction and faster convergence in downstream video generation.
(b) Our Dual-Stream Looped Diffusion Transformer refines RGB and geometry representations through shared-weight recurrence, achieving competitive generation quality with fewer backbone parameters.}
\label{fig:teaser}

\end{minipage}

    \par\vspace{6pt}
}]
\makeatother
\begin{abstract}
Predicting the outcomes of robotic manipulation requires modeling both visual appearance and evolving 3D geometry. This poses two challenges: video compression can obscure fine-grained robot motion, while jointly predicting RGB and geometry demands sufficient generative capacity. We introduce Loop4D, a framework that addresses these challenges through motion-preserving 4D representations and recurrent generative refinement. Our 4D VAE encodes RGB and pointmap sequences into compact spatiotemporal latents while preserving motion details and geometric structure. Our RGB VAE combines geometry-guided supervision with predictive reconstruction to reduce compression-induced motion artifacts. Our geometry VAE uses lightweight adapters to adapt pretrained RGB representations while preserving their priors. Building on these representations, our Dual-Stream Looped Diffusion Transformer maintains separate appearance and geometry streams with shared Transformer weights and cross-modal connections. By reusing Transformer blocks across loops, we increase effective depth without replicating backbone parameters, enabling models with fewer parameters to improve joint RGB–geometry prediction through additional computation. Experiments demonstrate high-quality 4D generation and validate Loop4D in simulated robotic manipulation tasks.
\end{abstract}

\section{Introduction}
\label{sec:intro}

Latent generative models~\cite{rombach2022ldm,peebles2023dit,blattmann2023align,zeng2022lion,zhang2023shape2vecset} have made visual generation more scalable by separating compact representation learning from generative prediction.
VAEs in image and video~\cite{zhao2026predictive,jiang2025geo4d,lu2026gae,tong2026videomaetok,kim2026grace,zhao2024cvvae,leng2025repae,guo2026vrae,chen2025dcae15,chen2025dcae} reduce the dimensionality of observations, while geometric latent spaces support generation beyond appearance alone.
For embodied world modeling~\cite{du2023unipi,zhu2025irasim,wu2024gr1,liang2024dreamitate,kim2026cosmospolicy,worldvla2025,qwenrobot2026,vjepa22025}, these representations must also preserve the robot and scene structure needed for interaction.
General-purpose video models provide a strong starting point, but their compression objectives and computational budgets are not tailored to these requirements.
This raises two questions: \emph{how should pretrained video representations be adapted to embodied 4D dynamics, and how can prediction improve within a fixed parameter budget?}

We begin with the strengths and limitations of general video VAEs~\cite{wan2025,wan22,ltx23}.
Their pretrained representations offer both compact compression and reusable visual priors.
However, compression can introduce artifacts around robotic arms, and RGB reconstruction objectives do not explicitly supervise metric geometry.
We find that preserving pretrained RGB priors is essential when adapting a VAE to geometry.
This observation motivates selective adaptation: retain pretrained visual priors while learning the geometric representations and predictive capabilities required by embodied tasks.
Existing 4D generation models~\cite{wonder3d,4dnex2025,mi2026one4d,fang2026worldreel,sv4d22025,l4gm2024,eg4d2024,dreamgaussian4d2023} often rely on large backbones, motivating us to improve generation quality with fewer parameters.
Building on our RGB and geometry latents, we introduce a Dual-Stream Looped Diffusion Transformer that increases effective depth through shared-weight recurrence while preserving cross-modal interactions.
This design achieves generation quality comparable to larger models without increasing backbone parameters.

Guided by this analysis, we introduce \textbf{Loop4D}, combining a \textbf{4D VAE tailored to embodied dynamics} with a \textbf{Dual-Stream Looped Diffusion Transformer}.
Our 4D VAE compresses appearance and geometry into compact spatiotemporal latents.
The RGB VAE combines geometry-guided supervision with reconstruction and prediction objectives~\cite{zhao2026predictive} to reduce compression-induced robot-arm artifacts.
The geometry VAE uses residual adapters and geometric constraints to adapt our pretrained RGB VAE while preserving its pretrained backbone and visual priors.

Building on these representations, our Dual-Stream Looped Diffusion Transformer maintains separate RGB and XYZ hidden states with shared Transformer weights and cross-modal connections.
However, directly applying existing looping strategies~\cite{chng2026looped,li2026skip,mozer2026recirculation} to our 4D generation setting does not reliably improve prediction quality.
Additional loops can degrade robot-region predictions, while appearance and geometry may require different update magnitudes.
We therefore introduce two adaptations.
First, per-loop prediction supervision and a robot-region non-degradation constraint encourage successive loops to refine manipulation-relevant regions.
Second, a lightweight shared gate uses the current hidden states and candidate increments to adapt update magnitudes separately for RGB and XYZ.
The model repeatedly applies shared middle layers within each denoising call, increasing effective depth without replicating backbone parameters.
These adaptations enable models with fewer parameters to use additional computation to approach the generation quality of larger models.
Our contributions are summarized as follows:
\begin{itemize}
    \item We introduce Loop4D, a framework that combines a high-fidelity 4D VAE tailored to embodied dynamics with a recurrent generative backbone.
    \item We develop a 4D VAE that combines geometry-guided predictive reconstruction for preserving robot motion with adapter-based geometry adaptation that retains pretrained RGB priors.
    \item We propose a Dual-Stream Looped Diffusion Transformer with robot-region refinement constraints and adaptive modality updates, improving generation through recurrent computation without replicating parameters.
\end{itemize}

\section{Related Work}
\label{sec:related_work}

\subsection{Generative Latent Representations}
\label{sec:latent_representations}

Latent diffusion models scale image and video generation through compact representations~\cite{ldm,dit,videoldm,cogvideox}. Beyond reconstruction fidelity, representation alignment and latent regularization can improve downstream generation~\cite{vavae,repa,eqvae2025,repae2025,diffusability2025}. Video autoencoders also explore compatibility with pretrained generators, generation-aware compression, and predictive learning~\cite{cvvae2024,grace2026,zhao2026predictive}. Geometric codecs and geometry-grounded video representations extend these ideas to point maps, motion, and 3D-consistent generation~\cite{geometrycrafter2025,motioncrafter2026,gae2026,vgga2026}. These studies motivate evaluating latents through both reconstruction and generation. Loop4D combines predictive RGB reconstruction and geometry-guided robot-region supervision with residual XYZ adaptation that preserves pretrained RGB priors.

\subsection{Embodied 4D World Modeling}
\label{sec:embodied_world_modeling}

Dynamic scene methods reconstruct observed motion with factorized fields or deformable Gaussians~\cite{hexplane,kplanes,fourDgs,deformgs}, while generative approaches synthesize content across time and viewpoints~\cite{fourdiffusion,cat4d,4dnex2025}. Embodied world models predict future observations conditioned on instructions or actions~\cite{unisim,robodreamer,tesseract,orv}. For manipulation, these predictions must preserve the robot's appearance and the spatial structure needed for interaction. Robot4DGen~\cite{robot4dgen2026} generates geometrically consistent RGB-D futures for manipulation, whereas PointWorld~\cite{pointworld} models action-conditioned 3D motion. Loop4D focuses on future RGB--XYZ generation from an initial observation and text instruction, studying task-adapted codecs and recurrent prediction.

\subsection{Looped Generative Computation}
\label{sec:looped_computation}

Shared-layer recurrence~\cite{dehghani2019universal, yang2024looped,geiping2025recurrent,lee2015deeply,ma2025inference,li2025reflectdit} increases the effective depth while reusing parameters~\cite{universaltransformer2018,recurrentdepth2025}. In diffusion models, inference-time scaling also includes noise search and feedback-based refinement~\cite{diffusioninferencescaling2025,reflectdit2025}. Looped-DiT~\cite{chng2026looped} repeats Transformer blocks within each denoising call, using intermediate supervision to support iterative refinement. Unlike additional sampling steps, these loops refine hidden states at a fixed noise level. Loop4D extends this design to coupled RGB and XYZ streams, introducing robot-region non-degradation constraints and adaptive modality updates. We examine how these mechanisms coordinate appearance and geometry, and evaluate the trade-off between generation quality and recurrent computation at a fixed backbone parameter count.

\section{Method}
\label{sec:method}

\begin{figure}[t]
    \centering
    \includegraphics[width=\columnwidth]{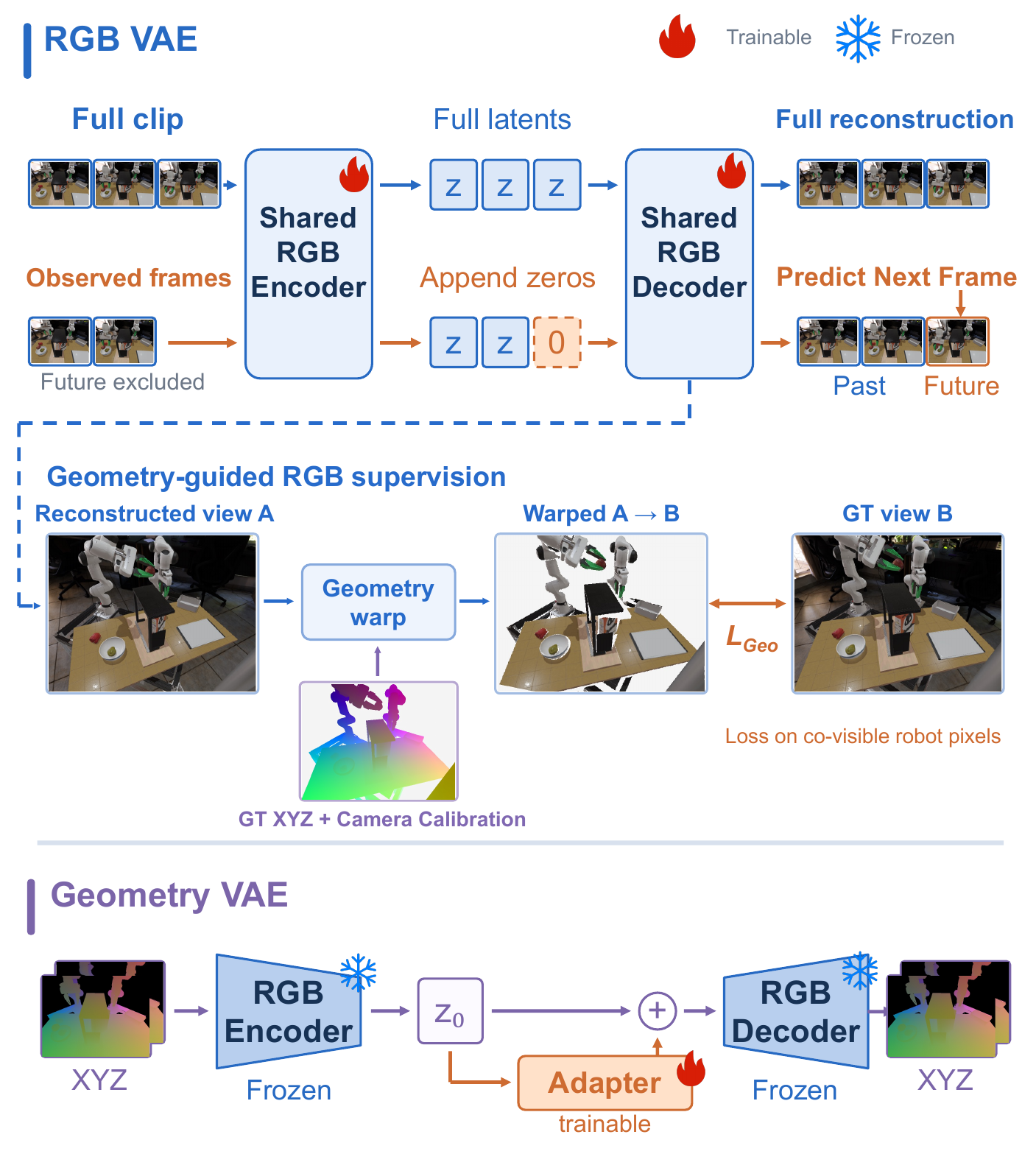}
    \caption{RGB VAE training uses predictive reconstruction and geometry-guided supervision. Geometry adaptation preserves pretrained RGB priors.}
    \label{fig:4d_vae_pipeline}
\end{figure}

Given an initial RGB-D observation and a text instruction, Loop4D predicts future RGB frames $I\in\mathbb{R}^{T\times H\times W\times3}$ and aligned camera-space XYZ maps $G$ of the same shape~\cite{robot4dgen2026}.
As shown in Fig.~\ref{fig:4d_vae_pipeline}, our 4D VAE uses RGB $(E_I,D_I)$ and geometry $(E_G,D_G)$ codecs to produce aligned spatiotemporal latents $z^I$ and $z^G$.
We first train the RGB VAE, then initialize and adapt the geometry VAE while preserving its RGB priors.
With both VAEs frozen, we train a Dual-Stream Looped Diffusion Transformer to jointly generate the two latent sequences for decoding.

\subsection{Loop4D: RGB VAE}
\label{sec:4d_vae}

\paragraph{Predictive Reconstruction.}
Spatiotemporal compression can weaken temporal continuity, particularly in robot regions undergoing large motions.
We combine predictive learning with reconstruction~\cite{zhao2026predictive} to preserve visual detail and capture temporal evolution.
Given $I=[I_{\mathrm{obs}},I_{\mathrm{future}}]$, we reconstruct the full clip from its observed prefix:
\begin{equation}
    \hat I^{\mathrm{pred}}=D_I\!\left([E_I(I_{\mathrm{obs}}),\mathbf{0}]\right).
    \label{eq:vae_prediction}
\end{equation}
Zero padding replaces withheld latent positions, enabling the same decoder to predict future frames without exposing them to the encoder.
We retain full-clip reconstruction and sample prediction steps with probability $p$:
\begin{equation}
    \mathcal L_{\mathrm{RGB}}=(1-p)\mathcal L_{\mathrm{rec}}+p\mathcal L_{\mathrm{pred}}.
    \label{eq:rgb_losses}
\end{equation}
Here, $\mathcal L_{\mathrm{rec}}$ combines full-clip $L_1$, mask-normalized robot-region supervision, and geometric warping.
$\mathcal L_{\mathrm{pred}}$ separately supervises observed and future frames, emphasizing future robot regions and matching robot-region temporal RGB differences, including the observed--future transition.
See the supplementary material for details.

\noindent\textbf{Geometry-guided RGB supervision.} Large robot motions can lead to overlapping arm artifacts and inconsistent appearance after VAE compression.
Since our 4D VAE models both appearance and geometry, we exploit its naturally available geometric information to further supervise RGB reconstruction in robot regions, encouraging consistent arm appearance during motion.
For synchronized views $a$ and $b$ at time $t$, we first locate the source pixel corresponding to a target pixel $q$:
\begin{equation}
\begin{aligned}
    X_t^b(q) &= \operatorname{BackProj}_b(q,d_t^b(q)),\\
    p_t(q) &= \operatorname{Proj}_a(T_{b\to a}X_t^b(q)).
\end{aligned}
    \label{eq:rgb_correspondence}
\end{equation}
Here, $\operatorname{BackProj}_b$ lifts a pixel to 3D using ground-truth depth $d_t^b$, $T_{b\to a}$ transforms it into camera $a$, and $\operatorname{Proj}_a$ projects it onto the source image; all use known camera calibration.
We then sample the reconstructed source RGB at this location:
\begin{equation}
    \hat I_t^{a\to b}(q)=\hat I_t^{a,\mathrm{rec}}(p_t(q)).
    \label{eq:rgb_warp_sampling}
\end{equation}
We compute $\ell_{\mathrm{warp}}$ as the mean absolute RGB error between the warped reconstruction and the observed target image over valid robot correspondences.
Only pixels labeled as robot in both views and satisfying depth-based visibility checks are included.
Applied only during reconstruction, this supervision complements predictive learning to address appearance artifacts in moving robot regions.
\subsection{Loop4D: Geometry VAE}
\label{sec:geometry_vae}

Pretrained RGB VAEs provide visual priors that we aim to retain when learning geometry.
Rescaling XYZ maps to the RGB input range $[-1,1]$ makes them compatible with the input interface, but does not eliminate the statistical differences between geometry and appearance.
Unrestricted fine-tuning may weaken these priors, motivating lightweight adaptation that preserves the RGB backbone while learning geometric corrections.

\noindent\textbf{Adaptation with preserved RGB priors.} We initialize the geometry codec from the trained RGB VAE and freeze its encoder and decoder to retain the inherited priors.
For normalized XYZ maps $\bar G$, let $z_0^G=E_G^0(\bar G)$ be the frozen encoder output.
A lightweight residual adapter produces
\begin{equation}
    z^G=z_0^G+\alpha A_\phi(z_0^G),
    \label{eq:geometry_adapter}
\end{equation}
which is decoded as $\hat{\bar G}=D_G^0(z^G)$.
Zero-initializing the adapter's output layer preserves the inherited mapping at initialization.
We further constrain latent changes with $\mathcal L_{\mathrm{prior}}=\|z^G-z_0^G\|_1$, keeping $z_0^G$ fixed.

\noindent\textbf{Robust geometric supervision.} To reduce sensitivity to depth outliers, we complement squared coordinate error with a Huber term.
We also supervise surface normals to constrain local orientation and discourage surface distortions.
Following point-map reconstruction objectives~\cite{motioncrafter2026}, we optimize
\begin{equation}
    \mathcal L_G=\mathcal L_{\mathrm{xyz}}+\lambda_n\mathcal L_{\mathrm{normal}}+\mathcal L_{\mathrm{aux}}+\lambda_z\mathcal L_{\mathrm{prior}}.
    \label{eq:geometry_loss}
\end{equation}
Here, $\mathcal L_{\mathrm{xyz}}$ combines squared error and Huber loss on normalized XYZ.
$\mathcal L_{\mathrm{normal}}$ measures cosine disagreement between normals computed from metric XYZ via spatial finite differences.
$\mathcal L_{\mathrm{aux}}$ combines weighted $L_1$ losses on metric depth, multiscale patch-centered depth, and spatial XYZ gradients.
All geometric terms require valid support.
We use encoder means and retain the original latent normalization without an additional KL penalty.

\subsection{Dual-Stream Looped Diffusion Transformer}
\label{sec:looped_dit}

\begin{figure*}[t]
    \centering
    \includegraphics[width=\textwidth]{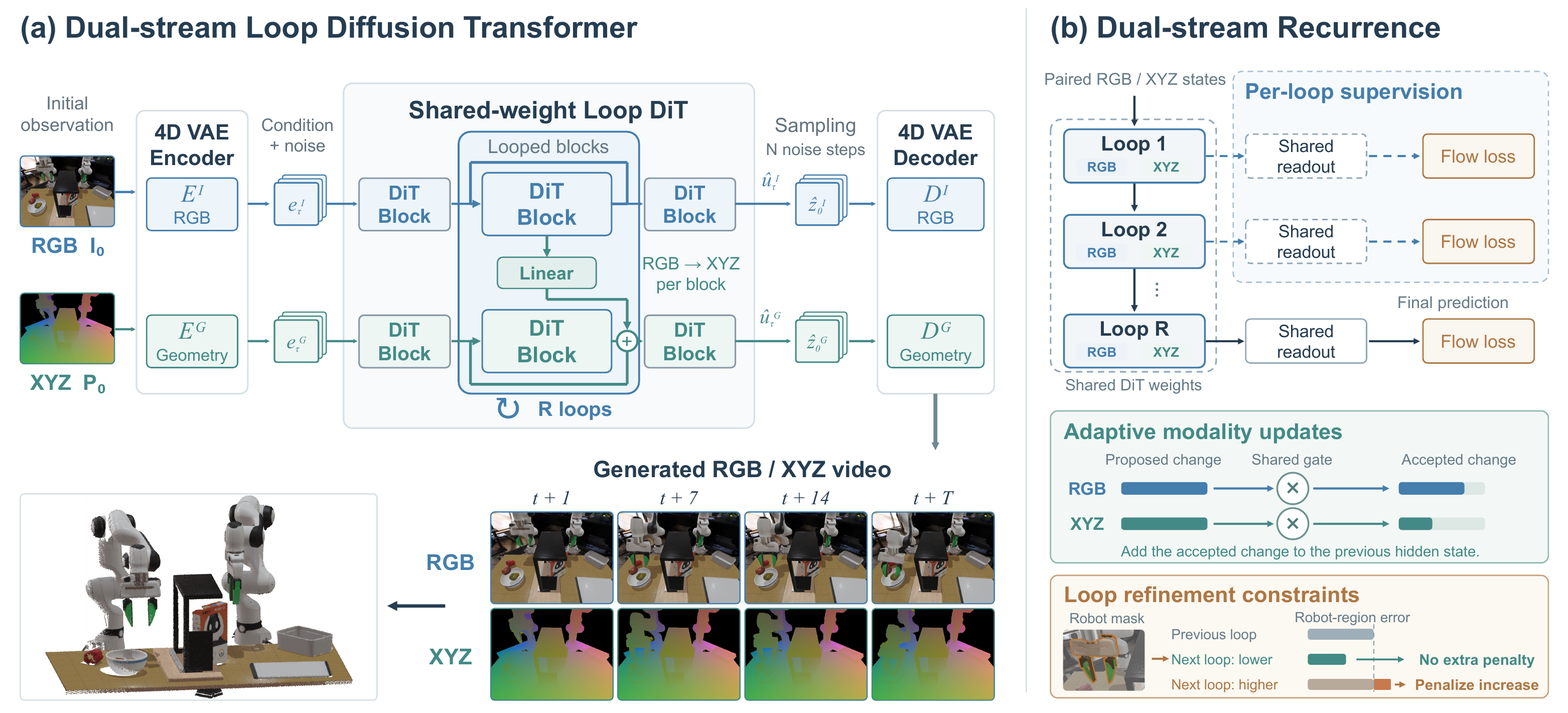}
\caption{\textbf{Overview of Loop4D.}
(a) A 4D VAE and Dual-Stream Looped Diffusion Transformer jointly generate RGB--pointmap videos.
(b) Shared-weight recurrence combines per-loop supervision, robot-region constraints, and adaptive modality updates.}
    \label{fig:overview}
\end{figure*}

Compact models require less memory for their weights, making them attractive for practical embodied applications.
We use shared-layer recurrence to increase effective depth at a fixed parameter count, allowing iterative refinement of internal representations without explicit reasoning tokens.
For 4D generation, this refinement must also coordinate RGB appearance and XYZ geometry, as both describe the same evolving scene.
We therefore introduce a \textbf{Dual-Stream Looped Diffusion Transformer} that couples modality-specific hidden states throughout the recurrent computation as shown in Fig~\ref{fig:overview}.

\noindent\textbf{Dual-stream Recurrence.} RGB and XYZ maintain separate hidden states while sharing Transformer weights.
Each shared block $F_l$ processes both streams, and a learned linear connection $B_l$ injects RGB features into the geometry stream:
\begin{equation}
\begin{aligned}
    h_{l+1}^I &= F_l(h_l^I;c),\\
    h_{l+1}^G &= F_l(h_l^G;c)
    +B_l(\operatorname{sg}(h_{l+1}^I)),
\end{aligned}
\label{eq:dual_stream}
\end{equation}
where $c$ denotes noise-time and text conditioning, and $\operatorname{sg}$ stops gradients along the lateral connection. \\
We retain all pretrained blocks and partition them into a prefix $P$, a recurrent middle $C_\theta$, and a suffix $S$.
Let $e_\tau$ denote the embedded noisy RGB and geometry latents, and $h_r$ the paired hidden states after round $r$:
\begin{equation}
\begin{aligned}
    h_0 &= P(e_\tau;c),\\
    h_r &= C_\theta(h_{r-1};c),\quad r=1,\ldots,R,\\
    \hat u_r &= \mathcal H(S(h_r;c)),
\end{aligned}
\label{eq:loop}
\end{equation}
where $\mathcal H$ is the shared prediction head.
Each round reuses the same middle blocks and RGB-to-XYZ connections, passing its hidden states directly to the next round.
The suffix and prediction head provide per-round supervision without feeding their outputs back into the recurrence.
All rounds occur within one denoising call, with the noise time, text, and initial-frame condition held fixed.
We also apply parameter-free Exclusive Self Attention (XSA)~\cite{zhai2026exclusive} within the recurrent layers.

\noindent\textbf{Loop refinement constraints. }To adapt looped refinement to embodied tasks, we explicitly constrain RGB and geometry predictions in robot regions.
For each sample and modality $m\in\{I,G\}$, we penalize increases in robot-region error between consecutive loops:
\begin{equation}
    \ell_{\mathrm{robot}}^{r,m}=\max\!\left(0,E_r^m-\operatorname{sg}(E_{r-1}^m)\right),
    \label{eq:robot_loss}
\end{equation}
where $E_r^m$ is the robot-region flow-matching error at loop $r$, and $\operatorname{sg}$ stops gradients through the preceding error.
We aggregate these penalties across samples, modalities, and loops, and combine them with per-loop flow-matching supervision.
This discourages additional loops from degrading robot-region predictions.
Implementation details are provided in the supplementary material.

\noindent\textbf{Adaptive modality updates.} RGB and pointmaps need not benefit from equal update magnitudes during recurrence.
We therefore adapt each stream's update to its current state:
\begin{equation}
    h_r^m=h_{r-1}^m+g_r^m\Delta h_r^m,
    \label{eq:adaptive_loop_gate}
\end{equation}
where $\Delta h_r^m=\tilde h_r^m-h_{r-1}^m$ is the candidate increment for modality $m\in\{I,G\}$.
From the second loop onward, a lightweight network predicts a scalar $g_r^m\in(0,1)$ for each sample and modality using pooled summaries of its current state and candidate increment.
The first loop is unchanged ($g_1^m=1$).
The gate is shared across modalities and loops and learned through the existing prediction objectives without additional annotations.
It adds a small, fixed parameter overhead that does not grow with the loop count.

\section{Experiments}
\label{sec:experiments}

We evaluate Loop4D in three main settings: (i) RGB and geometry reconstruction with the Loop4D VAE; (ii) future 4D video generation on simulated and real-world tasks; and (iii) simulated grasping using generated videos, following Robot4DGen~\cite{robot4dgen}.
We further examine how VAE adaptation affects generation and how looped refinement affects quality and computation.

\subsection{Experimental Setup}
\label{sec:exp_setup}

\noindent\textbf{Dataset.} We use the Robot4DGen dataset~\cite{robot4dgen}, which contains multi-view RGB-D demonstrations of three simulated and four real-world robot manipulation tasks.
We generate videos at a resolution of $640\times480$.

\noindent\textbf{Metrics.} We report rFVD~\cite{unterthiner2018fvd}, SSIM, and PSNR (dB) for RGB reconstruction.
For video generation, we report gFVD and gFID~\cite{heusel2017fid}, excluding the conditioning frame from generation metrics.
Geometry is evaluated using depth AbsRel and threshold accuracy $\delta_1$ ($1.25$).
Simulated grasping is evaluated by task success rate.

\noindent\textbf{Baselines.} For reconstruction, we compare Loop4D with MotionCrafter~\cite{motioncrafter2026}, HunyuanVideo~\cite{hunyuanvideo2024}, CogVideoX~\cite{cogvideox}, LTX-2.3~\cite{ltx23}, Wan2.1~\cite{wan2025}, Wan2.2~\cite{wan22}, and Robot4DGen VAEs~\cite{robot4dgen} .
For generation, we compare with video generators Wan2.1~\cite{wan2025}, Wan2.2~\cite{wan22}, and LTX-2.3~\cite{ltx23}, and 4D generators 4DNeX~\cite{4dnex2025} and Robot4DGen~\cite{robot4dgen}.
We additionally include 4D Gaussian ~\cite{som2025} as a reconstruction reference.
For simulated grasping, we compare with Diffusion Policy (DP)~\cite{chi2023diffusion}, and Robot4DGen~\cite{robot4dgen}.

\begin{table*}[t]
    \centering
    \caption{\textbf{Quantitative results of VAE reconstruction on Robot4DGen.}Our 4D VAE outperforms most mainstream VAEs.
}
    \label{tab:vae_reconstruction}
    \footnotesize
    \setlength{\tabcolsep}{4pt}
    \begin{tabular*}{\textwidth}{@{\extracolsep{\fill}}lccccc@{\hspace{12pt}}ccccc@{}}
        \toprule
        & \multicolumn{5}{c}{Simulation tasks}
        & \multicolumn{5}{c}{Real-world tasks} \\
        \cmidrule(lr){2-6}\cmidrule(lr){7-11}
        & \multicolumn{3}{c}{RGB}
        & \multicolumn{2}{c}{Geometry}
        & \multicolumn{3}{c}{RGB}
        & \multicolumn{2}{c}{Geometry} \\
        \cmidrule(lr){2-4}\cmidrule(lr){5-6}
        \cmidrule(lr){7-9}\cmidrule(lr){10-11}
        Method
        & rFVD$\downarrow$ & SSIM$\uparrow$ & PSNR$\uparrow$
        & AbsRel$\downarrow$ & $\delta_1\uparrow$
        & rFVD$\downarrow$ & SSIM$\uparrow$ & PSNR$\uparrow$
        & AbsRel$\downarrow$ & $\delta_1\uparrow$ \\
        \midrule
        MotionCrafter VAE~\cite{motioncrafter2026}
        & 81.93 & 0.9024 & 32.14 & 0.0143 & 99.36
        & 56.56 & 0.9498 & 33.59 & 0.0142 & 99.40 \\
        HunyuanVideo VAE~\cite{hunyuanvideo2024}
        & \textbf{15.45} & 0.9563 & 36.60 & \underline{0.0112} & 99.33
        & \textbf{11.06} & 0.9700 & 38.89 & 0.0121 & 99.02 \\
        CogVideoX VAE~\cite{cogvideox}
        & 18.28 & 0.9610 & 37.16 & 0.0232 & 99.05
        & 21.39 & \underline{0.9736} & 38.17 & 0.0256 & 98.55 \\
        LTX-2.3 VAE~\cite{ltx23}
        & 22.67 & 0.9437 & 34.78 & 0.0169 & 98.67
        & 27.31 & 0.9669 & 37.03 & 0.0189 & 97.91 \\
        Wan2.1 VAE~\cite{wan2025}
        & 21.47 & 0.9545 & 36.31 & 0.0238 & 99.11
        & 15.33 & 0.9711 & 38.24 & 0.0261 & 98.61 \\
        Wan2.2 VAE~\cite{wan22}
        & \underline{15.71} & \underline{0.9649} & \underline{37.34}
        & 0.0121 & 99.34
        & \underline{14.48} & 0.9703 & \underline{38.92}
        & 0.0122 & 99.01 \\
        Robot4DGen VAE~\cite{robot4dgen}
        & 72.72 & 0.9044 & 32.23 & 0.0114 & \textbf{99.47}
        & 109.2 & 0.9390 & 32.12 & \underline{0.0088} & \underline{99.81} \\
        \midrule
        Loop4D VAE (ours)
        & 22.77 & \textbf{0.9669} & \textbf{37.38}
        & \textbf{0.0109} & \underline{99.38}
        & 24.14 & \textbf{0.9773} & \textbf{39.38}
        & \textbf{0.0080} & \textbf{99.99} \\
        \bottomrule
    \end{tabular*}
\end{table*}

\subsection{Loop4D VAE Reconstruction Quality}
\label{sec:exp_reconstruction}
\begin{figure}[t]
    \centering
    \includegraphics[width=\columnwidth]{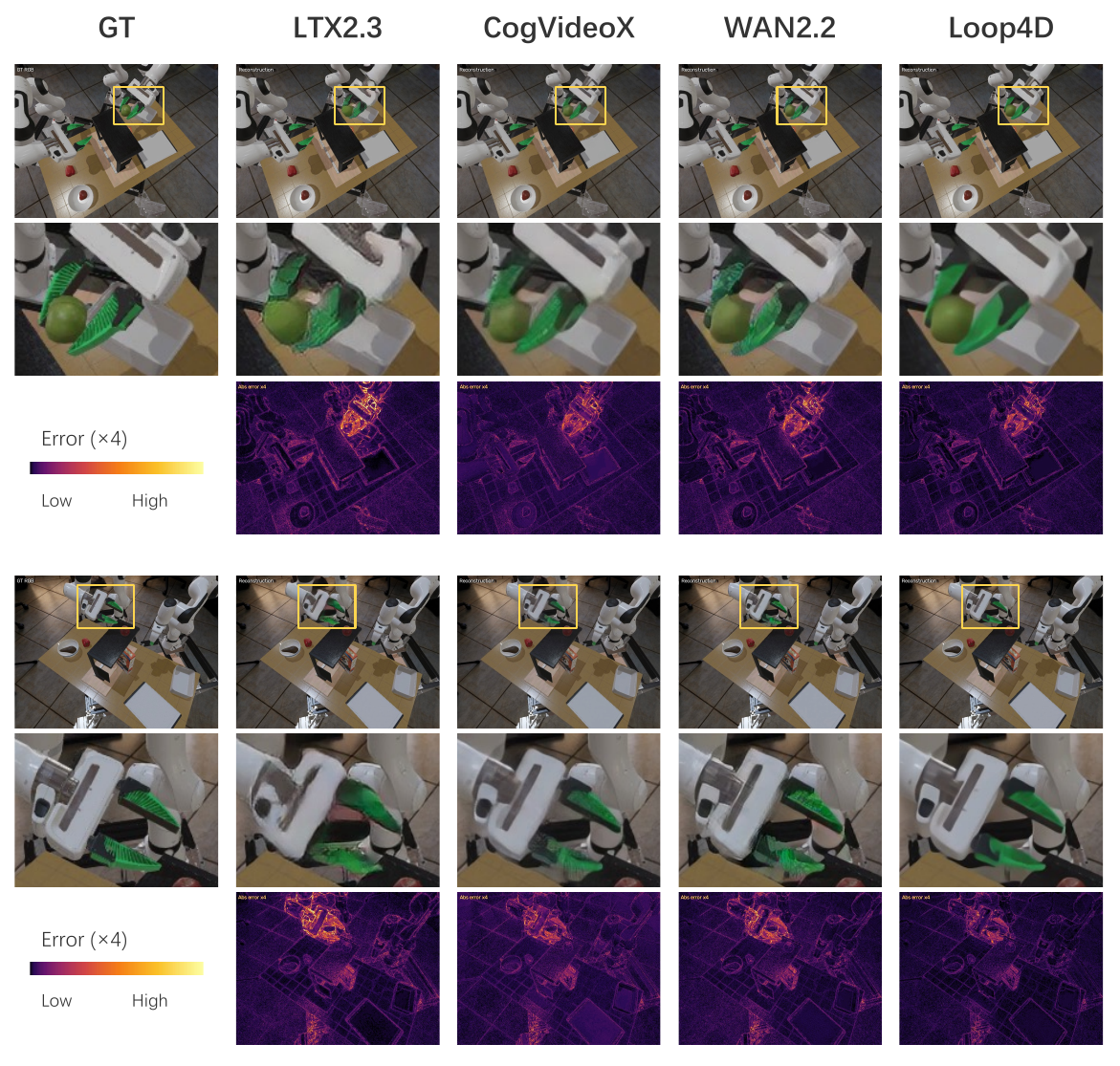}
    \caption{\textbf{Qualitative comparison of VAE reconstruction on Robot4DGen.}  }
    \label{fig:vae_reconstruction}
\end{figure}

\begin{figure*}[!t]
    \centering
    \includegraphics[width=\textwidth]{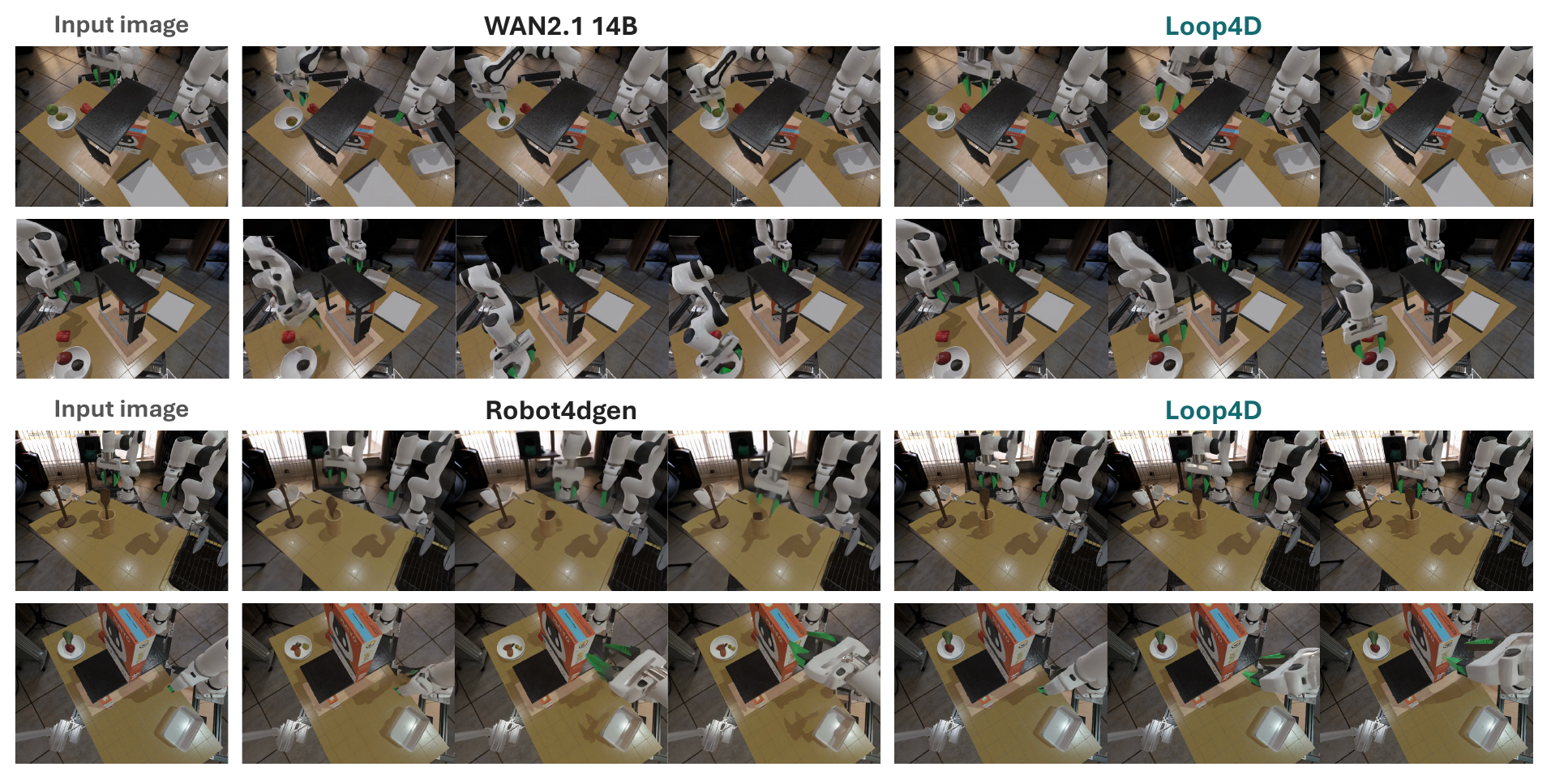}
    \caption{\textbf{RGB video generation with Loop4D on simulation tasks.}
    Loop4D maintains temporal consistency during robot motion and successfully generates robotic manipulation actions.}
    \label{fig:sim_rgb_generation}
\end{figure*}

Tab.~\ref{tab:vae_reconstruction} shows that Loop4D achieves the best SSIM, PSNR, and depth AbsRel in both domains, reducing AbsRel over Wan2.2 by 9.9\%/34.4\% in simulation/real-world tasks.
Real-world $\delta_1$ reaches 99.99\%.
Fig.~\ref{fig:vae_reconstruction} compares robot-arm structure and appearance during motion.

\subsection{Future 4D Video Generation}
\label{sec:exp_generation}

\begin{figure*}[t]
    \centering
    \includegraphics[width=\textwidth]{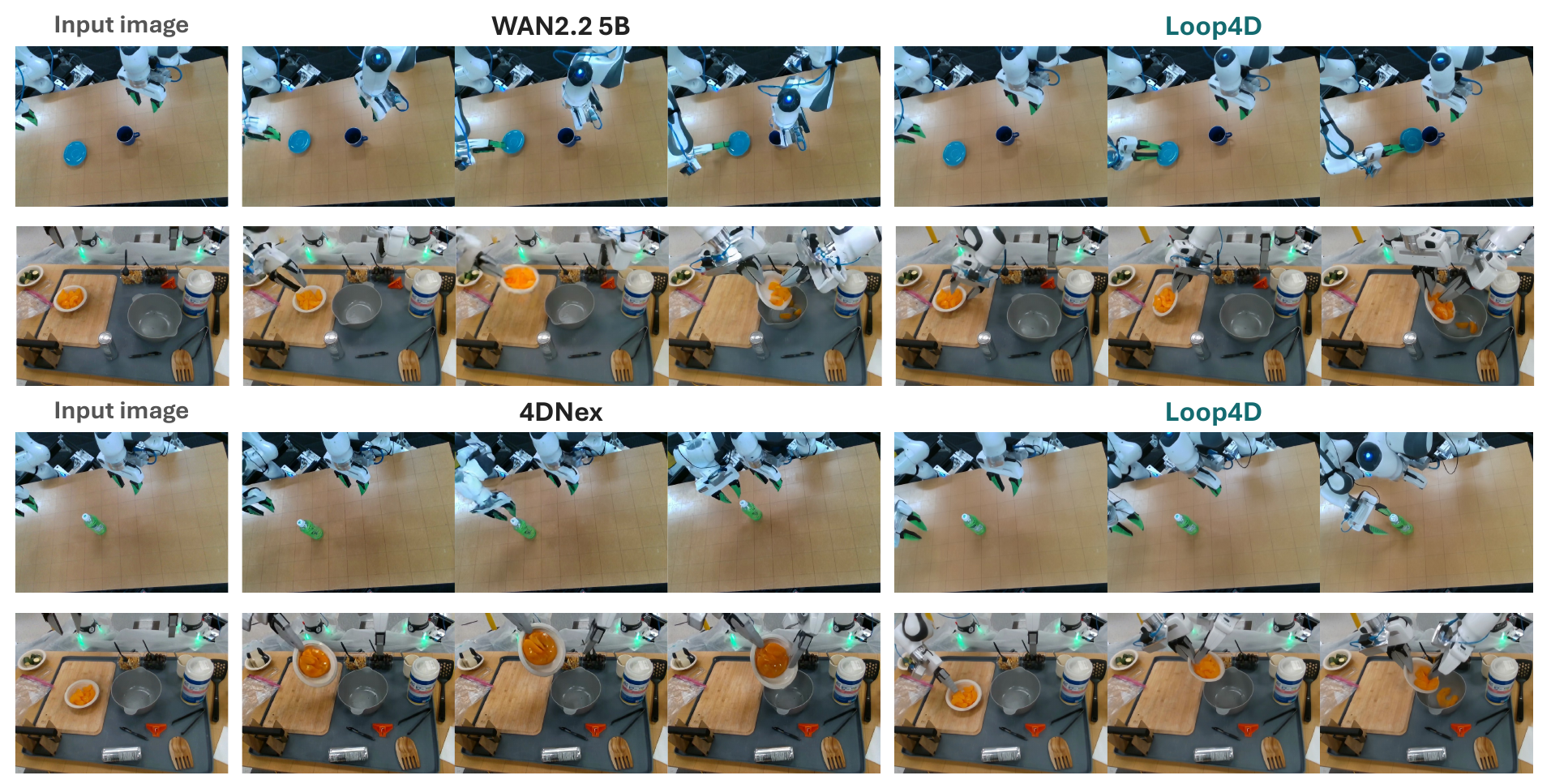}
\caption{\textbf{RGB video generation with Loop4D on real-world tasks.}
Loop4D generates realistic robot motions in real-world scenes.
The generated sequences depict successful completion of the target manipulation tasks.}
    \label{fig:real_rgb_generation}
\end{figure*}

\begin{figure}[t]
    \centering
    \includegraphics[width=\columnwidth]{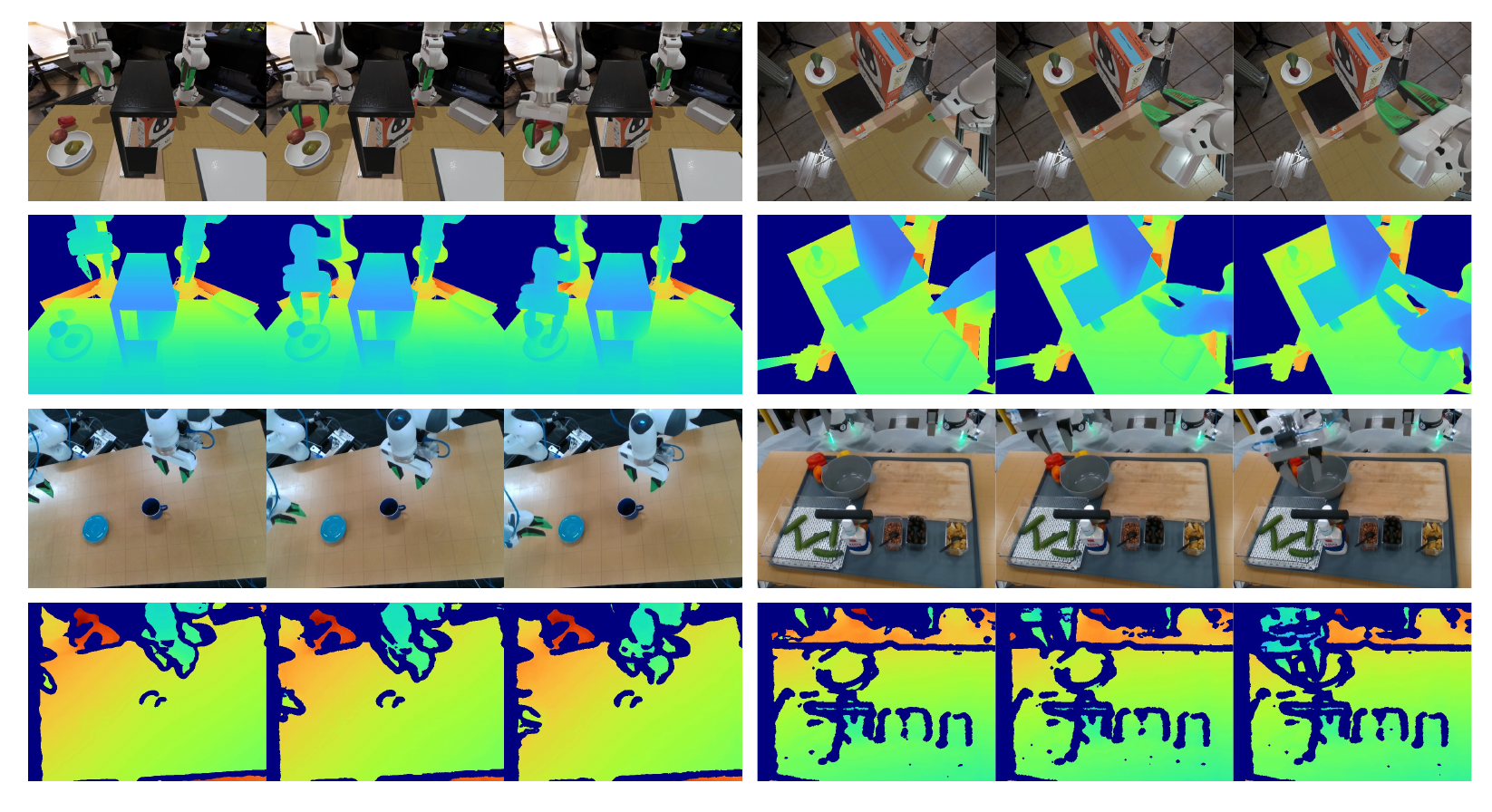}
    \caption{\textbf{4D video generation by Loop4D.}
    Generated RGB sequences and corresponding depth maps visualize appearance and geometry over time.}
    \label{fig:loop4d_rgb_depth}
\end{figure}

Tab.~\ref{tab:video_generation} compares Loop4D with RGB and 4D baselines.
Our 1.3B model achieves simulation gFVD/gFID of 449.33/49.37, outperforming 14B Wan2.1~\cite{wan2025} and 22B LTX-2.3~\cite{ltx23}.
Both variants outperform Robot4DGen in gFID and depth AbsRel across both domains.
The 5B model further improves all reported metrics over 1.3B, achieving the best reported gFID and AbsRel in both domains. Figs.~\ref{fig:sim_rgb_generation} and~\ref{fig:real_rgb_generation} show RGB comparisons with baselines, while Fig.~\ref{fig:loop4d_rgb_depth} presents our 4D generation results.
\begin{table*}[t]
    \centering
    \caption{\textbf{Quantitative comparison of 4D video generation on Robot4DGen.}
    We compare Loop4D with video generation models and 4D generation models, with all baselines fine-tuned on our dataset.
$\dagger$: reconstruction baseline without a comparable generative backbone.}
    \label{tab:video_generation}
    \footnotesize
    \setlength{\tabcolsep}{3pt}
    \begin{tabular*}{\textwidth}{@{\extracolsep{\fill}}lccccc@{\hspace{10pt}}cccc@{}}
        \toprule
        & & \multicolumn{4}{c}{Simulation tasks}
        & \multicolumn{4}{c}{Real-world tasks} \\
        \cmidrule(lr){3-6}\cmidrule(lr){7-10}
        Method & Params (M)
        & gFVD$\downarrow$ & gFID$\downarrow$
        & AbsRel$\downarrow$ & $\delta_1\uparrow$
        & gFVD$\downarrow$ & gFID$\downarrow$
        & AbsRel$\downarrow$ & $\delta_1\uparrow$ \\
        \midrule
        LTX-2.3 22B~\cite{ltx23} & 22,000
        & 502.82 & 60.46 & N/A & N/A
        & \underline{699.17} & 62.72 & N/A & N/A \\
        Wan2.1-14B~\cite{wan2025} & 14,000
        & 574.92 & 55.36 & N/A & N/A
        & \textbf{668.84} & \underline{47.65} & N/A & N/A \\
        Wan2.2-5B~\cite{wan22} & 5,000
        & 669.06 & 64.29 & N/A & N/A
        & 785.91 & 61.04 & N/A & N/A \\
        \midrule
        4D Gaussian$^\dagger$~\cite{som2025} & N/A
        & 1838.1 & 185.8 & 1.1266 & 10.93
        & 2543.3 & 201.3 & 0.6973 & 7.430 \\
        4DNeX~\cite{4dnex2025} & 14,000
        & 831.26 & 64.47 & 0.2352 & 68.07
        & 927.87 & 61.20 & 0.1676 & 75.58 \\
        Robot4DGen~\cite{robot4dgen} & \underline{2,400}
        & 682.44 & 91.70 & 0.1131 & 89.55
        & 918.23 & 134.2 & 0.1473 & 83.52 \\
        \midrule
        Loop4D-1.3B (ours) & \textbf{1,300}
        & \underline{449.33} & \underline{49.37}
        & \underline{0.0588} & \underline{93.67}
        & 738.77 & 53.89
        & \underline{0.0964} & \underline{92.73} \\
        Loop4D-5B (ours) & 5,000
        & \textbf{421.04} & \textbf{41.96}
        & \textbf{0.0483} & \textbf{94.74}
        & 721.24 & \textbf{41.63}
        & \textbf{0.0831} & \textbf{93.11} \\
        \bottomrule
    \end{tabular*}
\end{table*}

\subsection{Downstream Robot Manipulation}
\label{sec:exp_downstream}

Following Robot4DGen~\cite{robot4dgen}, we evaluate three tasks with 10 trials per method per task, using matched demonstrations and initializations.
Loop4D achieves higher average success than Diffusion Policy~\cite{chi2023diffusion} and Robot4DGen (Tab.~\ref{tab:sim_manipulation}).
Low video-based success may partly reflect trajectory-recovery and control errors in the shared execution pipeline.
\begin{table}[t]
    \centering
    \caption{\textbf{Simulated manipulation.} Task success rate (\%, $\uparrow$) on the three Robot4DGen tasks, with an unweighted task average.}
    \label{tab:sim_manipulation}
    \footnotesize
    \setlength{\tabcolsep}{3pt}
    \begin{tabular*}{\columnwidth}{@{\extracolsep{\fill}}lcccc@{}}
        \toprule
        Method & Cereal & Spatula & Apple & Avg. \\
        \midrule
        Diffusion Policy~\cite{chi2023diffusion} & 0.1 & 0.0 & 0.2 & 10.0 \\
        Robot4DGen~\cite{robot4dgen} & 0.3 & 0.1  & 0.1 & 16.7 \\
        Loop4D (ours) & \textbf{0.5} &\textbf{0.2} & \textbf{0.3} & \textbf{33.3} \\
        \bottomrule
    \end{tabular*}
\end{table}

\begin{table}[t]
    \centering
    \caption{\textbf{Ablation study of the RGB VAE.}}
    \label{tab:rgb_vae_ablation}
    \scriptsize
    \setlength{\tabcolsep}{1.5pt}
    \renewcommand{\arraystretch}{0.95}
    \begin{tabular*}{\columnwidth}{@{\extracolsep{\fill}}lccc@{}}
        \toprule
        \emph{(a) RGB reconstruction}
        & rFVD$\downarrow$ & SSIM$\uparrow$ & PSNR$\uparrow$ \\
        \midrule
        w/o Predictive Reconstruction & \textbf{19.01} & \textbf{0.9696} & \textbf{37.86} \\
        w/o Geometric Supervision & 24.61 & 0.9664 & 37.36 \\
        Loop4D VAE & 22.77 & 0.9669 & 37.38 \\
        \bottomrule
    \end{tabular*}
    \par\vspace{2pt}
    \begin{tabular*}{\columnwidth}{@{\extracolsep{\fill}}lccc@{}}
        \toprule
        \emph{(b) Downstream video generation}
        & FVD$\downarrow$ & FID$\downarrow$ & PSNR$\uparrow$ \\
        \midrule
        w/o Predictive Reconstruction & 697.5 & 75.03 & 19.88 \\
        w/o Geometric Supervision & 659.4 & 74.57 & 20.23 \\
        Loop4D VAE & \textbf{588.7} & \textbf{62.78} & \textbf{20.50} \\
        \bottomrule
    \end{tabular*}
\end{table}

\begin{table}[t]
    \centering
    \caption{\textbf{Comparison of different VAE fine-tuning strategies.}}
    \label{tab:vae_ablation}
    \scriptsize
    \setlength{\tabcolsep}{1.5pt}
    \renewcommand{\arraystretch}{0.95}
    \begin{tabular*}{\columnwidth}{@{\extracolsep{\fill}}lccccc@{}}
        \toprule
        \emph{(a) Reconstruction}
        & rFVD$\downarrow$ & PSNR$\uparrow$ & SSIM$\uparrow$
        & AbsRel$\downarrow$ & $\delta_1\uparrow$ \\
        \midrule
        Pretrained & \textbf{13.29} & 37.01 & 0.9613 & 0.0121 & 99.34 \\
        Decoder only & 15.28 & 38.25 & 0.9712 & 0.0167 & 99.16 \\
        Full fine-tuning & 14.12 & \textbf{38.49} & \textbf{0.9717} & 0.0129 & 99.24 \\
        Ours & 22.77 & 37.38 & 0.9669 & \textbf{0.0109} & \textbf{99.38} \\
        \bottomrule
    \end{tabular*}
    \par\vspace{2pt}
    \begin{tabular*}{\columnwidth}{@{\extracolsep{\fill}}lcccc@{}}
        \toprule
        \emph{(b) Generation}
        & gFVD$\downarrow$ & gFID$\downarrow$
        & AbsRel$\downarrow$ & $\delta_1\uparrow$ \\
        \midrule
        Pretrained & 742.5 & 76.20 & 0.0658 & 92.71 \\
        Decoder only & 682.9 & 67.27 & 0.0639 & 92.76 \\
        Full fine-tuning & 665.2 & 68.87 & \textbf{0.0526} & 93.85 \\
        Ours & \textbf{588.7} & \textbf{62.78} & \underline{0.0543} & \textbf{94.25} \\
        \bottomrule
    \end{tabular*}
\end{table}

\begin{figure}[t]
    \centering
    \includegraphics[width=\columnwidth]{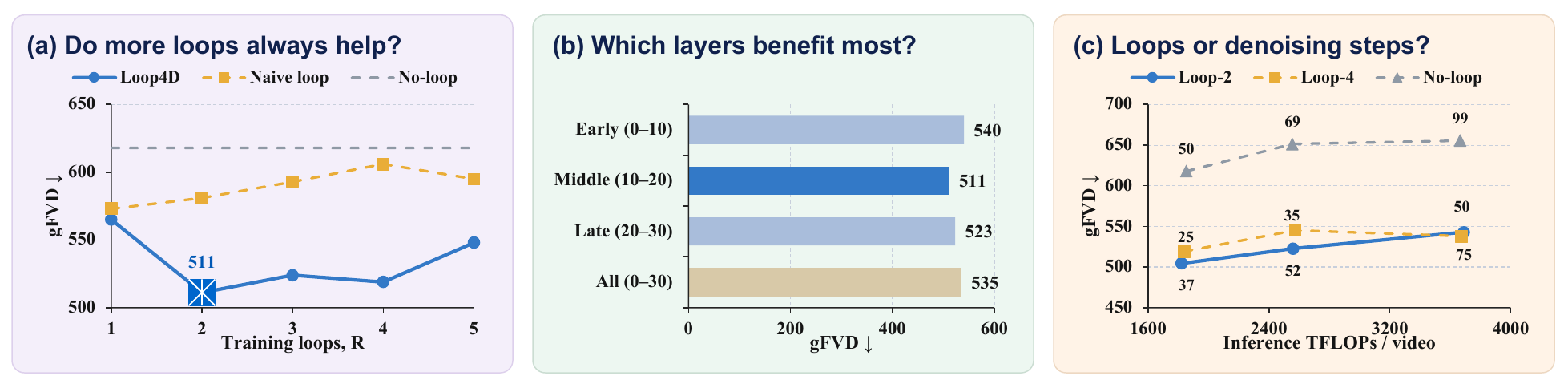}
    \caption{\textbf{Analysis of Dual-Stream LoopDiT.}
    (a) Effect of the number of loops.
    (b) Effect of looping different transformer layers.
    (c) Generation quality versus inference cost for different loop counts and sampling steps.
    Point labels in (c) indicate inference steps.}
    \label{fig:loop_analysis}
\end{figure}

\subsection{Ablation Studies}
\label{sec:exp_ablation}

\noindent\textbf{RGB VAE objectives.}
We ablate predictive learning and geometry-guided warping in Tab.~\ref{tab:rgb_vae_ablation}.
Although removing prediction improves reconstruction, it does not improve downstream generation, highlighting that reconstruction fidelity alone does not determine generative quality~\cite{yao2025vavae}.

\noindent\textbf{VAE fine-tuning strategy.} We compare pretrained codecs, decoder-only fine-tuning, full fine-tuning, and our adaptation using matched generators as shown in Tab.~\ref{tab:vae_ablation}.
Full fine-tuning yields higher reconstruction PSNR and SSIM, while our strategy achieves the lowest reported generation gFVD (588.7) and highest $\delta_1$ (94.25\%).
Full fine-tuning retains the lowest generation AbsRel (0.0526).
These results show that reconstruction fidelity alone does not determine downstream generation quality.
Despite its lower reconstruction PSNR and SSIM, Loop4D achieves better generation gFVD and $\delta_1$.
This highlights the importance of evaluating VAE through both reconstruction and downstream generation.
\begin{table}[t]
    \centering
\caption{\textbf{Loop Transformer ablations.}
(a) Generation quality and computational cost across model sizes.
(b) Effects of refinement constraints and adaptive modality updates.
All looped variants use $R=2$.
DiT training FLOPs include forward and backward passes per sample, excluding checkpoint recomputation; time reports 50-step inference denoising latency (s/video).}
    \label{tab:loop_ablation}
    \scriptsize
    \setlength{\tabcolsep}{1.5pt}
    \renewcommand{\arraystretch}{0.95}
    \begin{tabular*}{\columnwidth}{@{\extracolsep{\fill}}lcccc@{}}
        \toprule
        Variant
        & gFVD$\downarrow$
        & $\delta_1$ (\%)$\uparrow$
        & TFLOPs$\downarrow$
        & Time$\downarrow$ \\
        \midrule
        \multicolumn{5}{l}{\emph{(a) Model size and computation}} \\
        1.3B, non-looped & 631.435 & 93.88 & 182.64 & 36.92 \\
        1.3B, looped & 606.496 & 92.44 & 304.38	 & 49.93 \\
        5B, non-looped & 618.032 & 92.74 & 111.59	 & 21.84 \\
        5B, looped & \textbf{511.161} & \textbf{95.61} & 184.96 & 29.40 \\
        \midrule
        \multicolumn{5}{l}{\emph{(b) Refinement mechanisms (5B)}} \\
        Naive loop & 581.693 & 95.08 & 148.27 & 28.99 \\
        w/o Refinement Constraints & 545.781 & 94.66 & 184.96& 29.38 \\
        w/o Adaptive Modality Updates & 531.066 & 95.05 & 184.96 & 29.27 \\
        Full Loop4D & \textbf{511.161} & \textbf{95.61} & 184.96 & 29.40 \\
        \bottomrule
    \end{tabular*}
\end{table}

\noindent\textbf{Loop Transformer analysis.} With the 4D VAE fixed, we compare looped and non-looped 1.3B and 5B models as shown in Tab.~\ref{tab:loop_ablation}.
We report generation quality, parameter counts, and TFLOPs at fixed resolution, sequence length, and sampling steps to assess the trade-off between model size and recurrent computation.
We further compare direct looping with the full model and separately remove adaptive modality gating or the robot-region constraint to isolate their contributions to embodied prediction.

\subsection{Analysis of Dual-Stream LoopDiT}
\label{sec:loop_analysis}
We investigate three questions about recurrent computation in the 5B model.

\noindent\textbf{(a) Do more loops always help?}
Fig.~\ref{fig:loop_analysis}(a) compares Loop4D and naive looping with one to five loops.
Loop4D performs best at two loops, with no consistent gains thereafter, while naive looping generally degrades.
Increasing effective depth alone therefore does not guarantee better generation quality.
These results favor controlled refinement over unrestricted repetition.

\noindent\textbf{(b) Which layers benefit most from looping?}
Fig.~\ref{fig:loop_analysis}(b) compares early (0--10), middle (10--20), late (20--30), and full-backbone looping at $R=2$.
The first three settings repeat the same number of blocks, isolating the effect of layer placement.
Middle-layer looping achieves the best generation quality, outperforming full-backbone looping with less computation.
This suggests that where refinement occurs matters more than simply repeating more layers.

\noindent\textbf{(c) Can more denoising steps replace looping?} In Fig.~\ref{fig:loop_analysis}(c), we compare loop-2, loop-4, and no-loop settings.
For each setting, we fix the loop count and vary the denoising steps, plotting generation quality against total inference TFLOPs.
Loop-2 uses 37, 52, and 75 inference steps, while loop-4 and no-loop use 25/35/50 and 50/69/99 steps, respectively, giving approximately matched computational budgets.
Both looped settings outperform the non-looped model at comparable cost, suggesting that internal refinement uses computation more effectively than additional denoising steps within the evaluated budgets.

\section{Conclusion}
\label{sec:conclusion}

We presented Loop4D, combining a 4D VAE tailored to embodied dynamics with a Dual-Stream Looped Diffusion Transformer. The VAE preserves motion details and pretrained RGB priors, while Loop refinement constraints and adaptive modality updates guide recurrent refinement without replicating backbone parameters. Experiments demonstrate high-quality 4D generation and validate Loop4D in simulated manipulation, highlighting the value of adapting latent representations and generative computation.

\section*{LLM Usage Statement}
We used OpenAI ChatGPT and Codex to assist with drafting, translating, and editing manuscript text, including related-work prose and reference lookup; developing and running training, evaluation, and efficiency-measurement code; analyzing experimental results; and preparing figures and visualizations under the authors' direction.
All reported results, scientific claims, and conclusions were reviewed and verified by the authors, who take full responsibility for the submission.

\end{document}